\documentclass[letterpaper,10pt,conference]{ieeeconf}

\IEEEoverridecommandlockouts
\usepackage{amsmath}
\usepackage{amssymb}
\usepackage{array}
\usepackage{booktabs}
\usepackage{cite}
\usepackage{graphicx}
\usepackage{multirow}
\usepackage{url}
\usepackage{xcolor}

\newcommand{\method}{Human2Dex}
\newcommand{\gpir}{\textsc{GPIR}}

\ifdefined
\fi

\title{\LARGE\bfseries
Human2Dex: Dexterous Robot Learning from Wrist-Centric\\
Human Demonstrations
}

\author{ %
Jiachen Zhang$^{1}$,
Yuanyuan Li$^{1}$,
Shaqi Luo$^{1,*}$%
\thanks{%
$^{1}$Jiachen Zhang, Yuanyuan Li, and Shaqi Luo are with
the Beijing Academy of Artificial Intelligence, Beijing, China.%
}%
\thanks{%
$^{*}$Corresponding author: Shaqi Luo
(\texttt{sqluo@baai.ac.cn}).%
}%
}

\makeatletter

\newif\ifoverviewnumbered
\overviewnumberedfalse

\let\old@maketitle\@maketitle

\renewcommand{\@maketitle}{%
    \old@maketitle
    \vspace{-0.15in}
    \begin{center}
        \includegraphics[
            width=\textwidth,
            trim=10 140 0 120,
            clip
        ]{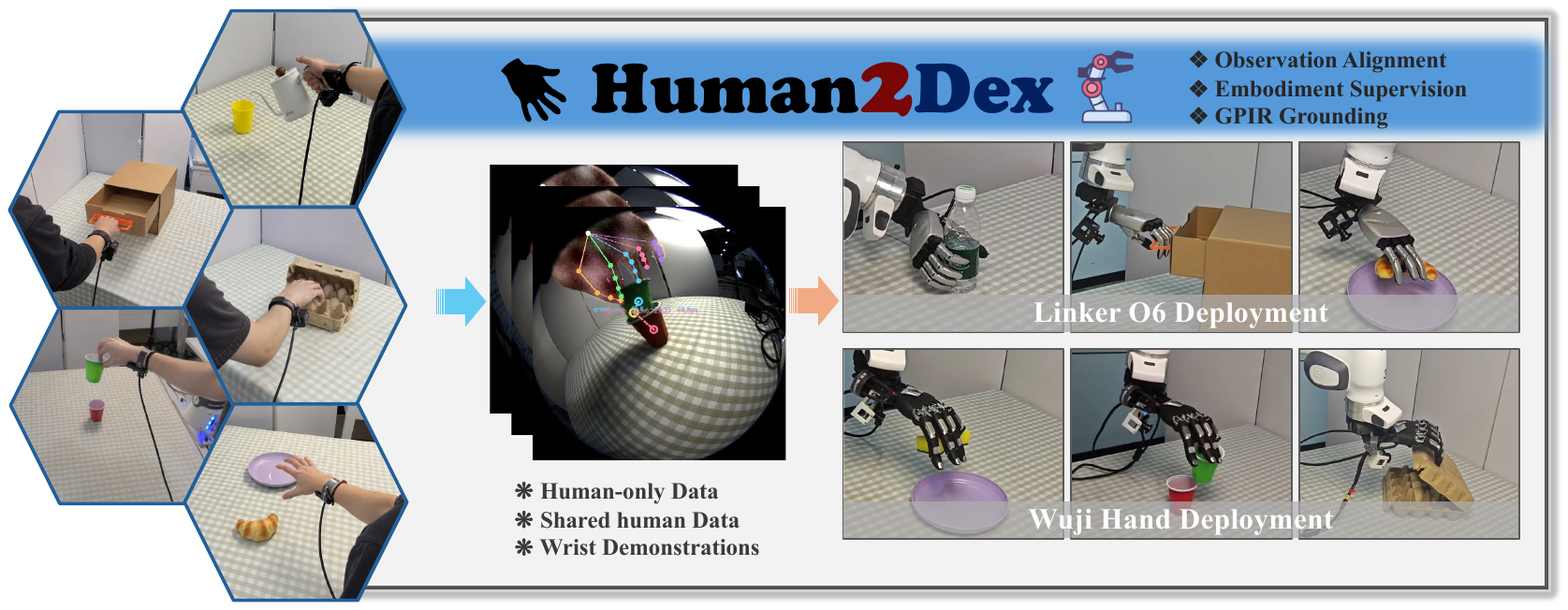}

        \vspace{2pt}

        {\small
        \ifoverviewnumbered
        \else
            \refstepcounter{figure}%
            \global\overviewnumberedtrue
            \label{fig:overview}%
        \fi

        \@makecaption{\fnum@figure}{%
        \textbf{Human2Dex overview.}
        A wrist-mounted camera records natural bare-hand demonstrations.
        The shared wrist-centric observation and 21-point hand interface
        are converted offline into separate executable training sets for
        the Linker O6 and Wuji hands. Visual adaptation and GPIR provide
        shared interface semantics, whereas retargeting, action labels,
        normalization, and diffusion policies remain hand--task specific.
        }
        }

    \end{center}
    \vspace{4pt}
}

\makeatother

\begin{document}

\maketitle

\thispagestyle{empty}
\pagestyle{empty}

\begin{abstract}

Collecting dexterous manipulation demonstrations for each robot hand requires repeated task-specific data acquisition. We ask whether one shared corpus of natural human demonstrations can instead serve as a reusable data interface for different dexterous hands, without collecting task demonstrations on each target robot. We present \method{}, a wrist-centric human demonstration interface that separates shared human observation and motion semantics from embodiment-specific execution. It aligns observations through local hand-appearance adaptation and a shared 21-point hand representation with consistent structural semantics across human and robot observations. It constructs executable supervision by retargeting fused human hand motion to each target hand, and introduces the Grasp-Pocket Interaction Representation (\gpir{}) to express tracked objects relative to a hand-centered Functional Grasp Pocket. The corpus is converted into separate training sets for Linker O6 and Wuji, while action labels, normalization, training, and diffusion-policy checkpoints remain hand-specific. Across six real-world tasks, \method{} achieves 109/120 (90.8\%) successes on Linker O6 and 97/120 (80.8\%) on Wuji. Across both hands, Human-Raw achieves 63/240 (26.3\%) successes, whereas the complete \method{} configuration achieves 206/240 (85.8\%). These results support the use of wrist-centric human demonstrations as a reusable interface across calibrated dexterous hands while retaining embodiment-specific policy learning. The project website is publicly available at \url{https://human2dex.pages.dev/}.

\end{abstract}
\section{Introduction}

Dexterous manipulation policies require demonstrations that jointly capture arm motion, high-DoF hand configuration, and contact-rich task progress. Acquiring such demonstrations separately for each target hand is costly and difficult to scale. Many existing systems make data collection more compatible with robot deployment through instrumented motion capture, robot-like interfaces, or wearable hardware~\cite{dexcap,umi,legato,dexumi,realdexumi,dexop}. These interfaces can reduce the correspondence burden during transfer, but they often establish robot compatibility partly through the acquisition setup, control interface, or embodiment-aware mapping. When the target embodiment changes, the corresponding conversion or embodiment-specific setup must still be established for the new hand. This motivates a higher-level question: how can a reusable human demonstration interface convert the same human manipulation episodes into executable training data for different dexterous hands, without collecting task demonstrations on each target robot?

We study this question from a wrist-centric perspective. Head-mounted egocentric video provides scalable human data and broad scene context~\cite{egomimic,egodex,humanego,vividex,unidex}, whereas a wrist-mounted camera provides local hand--object observations that more closely correspond to the eye-in-hand views used for close-range dexterous control. This similarity in observation arrangement makes wrist-centric acquisition a natural starting point for demonstration reuse. However, matching the observation arrangement does not by itself establish correspondence between human and robot data. The two settings can still differ in hand appearance, articulated structure, kinematics, command spaces, and grasp-related geometry.

Even under wrist-to-eye-in-hand alignment, three linked mismatches remain. First, human and robot hands differ in appearance and articulated structure, so the same visual content does not automatically provide the same structural semantics. Prior work has addressed this issue through appearance adaptation, robot-hand rendering, and structured hand representations~\cite{mirage,dexumi,dexpilot,anyteleop}. Second, human motion is not itself an executable command for a target hand: Linker O6 and Wuji have different kinematic structures, degrees of freedom, joint limits, and native action spaces. Human-to-robot retargeting and embodiment-aware motion mapping therefore remain necessary for constructing target-specific supervision~\cite{dexcap,dexpilot,dexh2r,mocaprobot}. Third, even after observation semantics are aligned and executable supervision is constructed, a policy must infer how the task object is positioned relative to the hand's current grasp configuration. Image-centered object coordinates indicate where an object appears, but do not directly express its relation to the closing hand. Existing work has explored hand--object and functional grasp representations for grasp synthesis, contact modeling, and perception~\cite{functionalgrasp,drograsp,humanego,contactpose,hot3d}. Thus, structural correspondence supports observation interpretation, executable supervision supports target-hand policy learning, and an explicit object--hand relation can provide a more direct interaction cue, particularly when grasp alignment matters.

We present \method{}, a reusable wrist-centric human demonstration interface that shares human demonstrations and their observation and motion semantics across embodiments while retaining target-hand-specific executable supervision and policy learning. To align observations, \method{} combines mask-guided local hand-appearance adaptation with a shared 21-point hand representation, establishing consistent structural semantics for human and robot observations. To construct executable supervision, fused human hand motion is retargeted separately to each calibrated target hand, producing native action labels for Linker O6 and Wuji. To represent interaction geometry, \method{} introduces the Grasp-Pocket Interaction Representation (\gpir{}), which expresses tracked task objects relative to a hand-centered Functional Grasp Pocket defined by the thumb--index grasp center. The same human demonstration corpus is therefore converted into separate training sets for the two target hands, while action labels, normalization statistics, policy training, and diffusion-policy checkpoints remain specific to each hand--task pair. No task demonstrations are collected with the target robot hands.

We evaluate whether this interface supports human-data reuse across embodiments under real-world dexterous manipulation. A matched Linker O6--Croissant study examines wrist- versus head-centric observation, while the main evaluation covers Linker O6 and Wuji across six real-world tasks. Additional reference-frame and task-wise analyses examine selected design choices and the task dependence of interaction grounding.

Our contributions are:
\begin{itemize}
\item \textbf{A reusable human demonstration interface for multiple dexterous embodiments.}
Human2Dex converts one wrist-centric human demonstration corpus into separate hand-specific training sets for Linker O6 and Wuji, allowing each hand to learn an independent policy without collecting task demonstrations on the target robots.

\item \textbf{A grasp-centered interaction representation for policy conditioning.}
The Grasp-Pocket Interaction Representation expresses tracked objects relative to a hand-centered Functional Grasp Pocket, complementing structural observation alignment and target-specific action supervision with an explicit object--hand relation.

\item \textbf{A two-hand, six-task physical evaluation with targeted design analyses.}
We evaluate the complete interface on two calibrated dexterous hands and six real-world tasks, complemented by controlled comparisons of observation viewpoint and object-reference coordinates and task-wise analysis of interaction-grounding behavior.
\end{itemize}

\section{Related Work}

\subsection{Learning Robot Policies from Human Demonstrations}

Human-data interfaces differ in where compatibility between human demonstrations and robot execution is established. Portable robot-like tools, motion-capture systems, teleoperation platforms, and wearable interfaces introduce deployment-relevant visual, kinematic, or contact signals during collection~\cite{mimicplay,umi,legato,dexcap,holodex,anyteleop,opentelevision,dexumi,realdexumi,dexop,dexvitac}. Recent UMI-style systems extend demonstration interfaces toward richer physical sensing and broader motion coordination. OmniUMI~\cite{omniumi} integrates visual, tactile, and force-related sensing within a handheld interface for contact-rich manipulation. BifrostUMI~\cite{bifrostumi} combines lightweight VR tracking with UMI-inspired grippers to collect robot-free demonstrations, from which a policy predicts human keypoint trajectories that are retargeted for humanoid whole-body execution. 

Other approaches preserve more natural human behavior through egocentric video, human play, or human--robot co-training, while some convert human data through simulation or limited robot data~\cite{egomimic,egodex,humanego,motiontrans,vividex,unidex,dexwild,robotictelekinesis,dextransfer,dexmimicgen}. Among the closest interface-based approaches, UMI establishes
deployment-compatible observations and actions through a portable
robot-like gripper, while DexUMI~\cite{dexumi} further reduces the human--robot
kinematic gap using a wearable hand exoskeleton together with robot-hand
visual adaptation. RealDexUMI~\cite{realdexumi} instead shares the dexterous end-effector
and sensing modules between demonstration collection and deployment,
thereby minimizing end-effector mismatch. HumanEgo~\cite{humanego} follows a different
route by lifting egocentric human video into an explicit entity-level
hand--object interaction representation. Human2Dex studies a complementary
setting: the source remains wrist-centric bare-hand manipulation, and
embodiment adaptation is performed after collection to construct native
action supervision and separate policies for dexterous hands with different
kinematics and action spaces.

\subsection{Visual and Kinematic Cross-Embodiment Alignment}

Cross-embodiment alignment involves both visual identity and articulated motion. Visual adaptation methods reduce differences between source and target hands through cross-painting, robot-hand rendering, or fine-grained hand--object segmentation~\cite{mirage,dexumi,egohos}. Hand estimators and teleoperation systems provide structured landmarks or mappings for transferring human motion under different morphologies and kinematic constraints~\cite{mediapipe,hamer,dexpilot,anyteleop,dexh2r,mocaprobot,crossdex}. Human2Dex uses a 21-point hand representation as an intermediate interface for shared observation semantics and motion fusion, while retaining embodiment-specific retargeting, action labels, and policy checkpoints for each target hand. Thus, the shared representation defines data semantics rather than a common robot action space.

\subsection{Object-Centric and Functional Grasp Representations}

Object-centric representations provide relational information beyond isolated hand or object states. FunctionalGrasp and D(R,O) Grasp use semantic hand--object structure or robot--object descriptions for grasp generation, while ContactPose and HOT3D support contact modeling and egocentric hand--object tracking~\cite{functionalgrasp,drograsp,contactpose,hot3d}. Human2Dex uses the Grasp-Pocket Interaction Representation (\gpir{}) for a narrower purpose: sequential wrist-view policy control. GPIR expresses tracked task objects relative to a hand-centered Functional Grasp Pocket and retains object scale and tracking reliability. It is therefore a low-dimensional grasp-relative interaction cue, rather than a general object reconstruction, universal grasp representation, or complete contact model.
\section{Method}
\label{sec:method}

\subsection{Shared Human Source and Conversion Interface}
\label{sec:shared-source}

Human2Dex converts one shared wrist-centric human demonstration source into
hand-specific training data for multiple dexterous embodiments. The
overall interface is illustrated in Fig.~\ref{fig:overview}. For task $\tau$,
the human demonstration corpus is
\begin{equation}
\mathcal{D}_{h,\tau}
=
\left\{
I_t,\,
T_t^w,\,
P_t^p,\,
P_t^v,\,
U_t^v
\right\}_{t=1}^{T_\tau},
\label{eq:human-source}
\end{equation}
where $I_t\in\mathbb{R}^{H\times W\times3}$ is the wrist-camera RGB image,
$T_t^w\in SE(3)$ is the human wrist pose,
$P_t^p,P_t^v\in\mathbb{R}^{21\times3}$ are 3-D hand structures from PICO
tracking and a wrist-RGB hand model, and
$U_t^v\in\mathbb{R}^{21\times2}$ contains direct image-space landmarks. The wrist-image model uses a DINO encoder shared by a 3-D MANO branch and a
direct 2-D keypoint branch. The 2-D landmarks support image-space rendering
and grasp-reference construction, whereas the 3-D structures support motion
fusion and target-hand retargeting.

For target hand $r$ and task $\tau$, Human2Dex constructs
\begin{equation}
\boxed{
\mathcal{D}_{r,\tau}
=
\Psi_{r,\tau}
\left(
\mathcal{D}_{h,\tau}
\right)
=
\left\{
\bar I_t,\,
q_t^r,\,
z_t,\,
a_t^r
\right\}_{t=1}^{T_\tau}
}
\label{eq:data-conversion}
\end{equation}
where $\bar I_t$ is the converted visual observation, $q_t^r$ is the
target-hand state, $z_t$ is the interaction state, and $a_t^r$ is the
executable arm--hand action. The conversion contains three complementary
branches: observation alignment, target-specific action supervision, and
grasp-centered interaction grounding. The same human source is reused for Linker O6 and Wuji. Retargeting is embodiment-specific, while action labels, normalization
statistics, policy training, and checkpoints are constructed separately for
each hand--task pair. Adding a new target hand therefore requires its
kinematic model, retargeter, native action conversion, normalization, and
policy training, but does not require recollecting the task demonstrations.
No task demonstrations are collected with the target robot hands.

\begin{figure}[t]
    \centering
    \includegraphics[
        width=\columnwidth,
        trim=100bp 145bp 90bp 140bp,
        clip
    ]{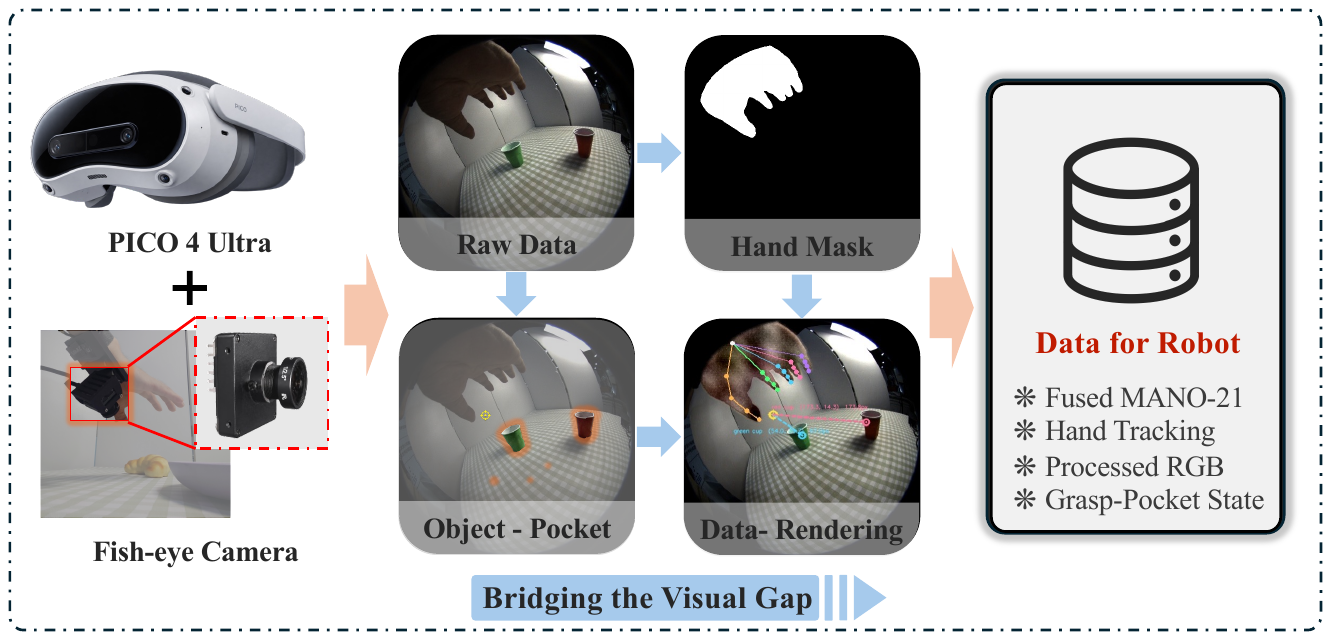}
    \caption{Human2Dex conversion interface. Wrist RGB, PICO tracking, and a
    wrist-image hand model provide measurements for structural observations,
    target-specific action supervision, and grasp-centered interaction
    grounding. The resulting data are converted offline for each target hand.}
    \label{fig:conversion}
\end{figure}

\subsection{Hand-Structure-Aligned Observations}
\label{sec:visual-observations}

The observation branch adapts the human-hand appearance and adds a shared
structural description. Given a hand mask $M_t^h$, local appearance
randomization is defined as
\begin{equation}
I_t^{\mathrm{aug}}
=
(1-M_t^h)\odot I_t
+
M_t^h\odot\mathcal{A}(I_t),
\label{eq:hand-randomization}
\end{equation}
where $\mathcal{A}$ is the appearance-randomization operator. The object and
background remain unchanged.

Let $\Gamma_t^h$ map the human image to the $224\times224$ policy image, and
let $\mathcal{R}$ render the 21-point hand structure:
\begin{equation}
\bar I_t^h
=
\mathcal{R}
\left(
\Gamma_t^h(I_t^{\mathrm{aug}}),
\Gamma_t^h(U_t^v)
\right).
\label{eq:human-rendering}
\end{equation}
A fixed topology and color assignment preserve landmark and finger identities
across observations.

At deployment, robot landmarks are generated from forward kinematics and
camera projection:
\begin{equation}
\bar U_{t,j}^r
=
\Gamma_t^r
\left[
\Pi
\left(
K_r,T_{C\leftarrow B,t}^r,F_j^r(q_t^r)
\right)
\right],
\label{eq:robot-landmarks}
\end{equation}
where $F_j^r(q_t^r)$ returns landmark $j$ in the robot base frame and
$K_r$ denotes the camera intrinsics. Since the camera is rigidly mounted
to the moving wrist, $T_{C\leftarrow B,t}^r$ is updated at each frame from
the current end-effector pose and the fixed hand--eye calibration. Human and
robot observations use the same landmark topology, colors, and image-space
coordinate convention.

RGB images, hand masks, object masks, and landmarks undergo the same crop,
resize, and image-space perturbation. The observation branch outputs the
registered structural observation $\bar I_t$.

\subsection{Target-Specific Action Supervision}
\label{sec:action-supervision}

The action branch converts the shared human motion into executable commands
for each target hand. The PICO and wrist-RGB 3-D structures are first
expressed in the same wrist-centered MANO frame. For landmark $j$, the two
estimates are combined using confidence-weighted interpolation:
\begin{equation}
\widehat P_{t,j}
=
\frac{
c_{t,j}^{p}P_{t,j}^{p}
+
c_{t,j}^{v}P_{t,j}^{v}
}{
c_{t,j}^{p}+c_{t,j}^{v}+\epsilon
},
\label{eq:motion-fusion}
\end{equation}
where $c_{t,j}^{p}$ and $c_{t,j}^{v}$ denote the confidence of the PICO and
wrist-RGB estimates, respectively. Invalid measurements receive zero weight.
The fused sequence is subsequently processed by an adaptive Kalman filter to
suppress temporal jitter and bridge short tracking dropouts, yielding the
fused 21-point structure $P_t$.

For each target hand, retargeting is formulated as a nonlinear kinematic
optimization over geometric hand relations rather than a direct mapping
between human and robot joint angles:
\begin{equation}
q_t^r
=
\rho_r(P_t;\mathcal{K}_r)
=
\underset{q\in\mathcal{Q}_r}{\arg\min}\,
\mathcal{E}_r(q;P_t),
\label{eq:retargeting}
\end{equation}
where $\mathcal{E}_r$ compares geometric features extracted from the fused
human landmarks with their robot counterparts computed through the forward
kinematics $\mathcal{K}_r$, and $\mathcal{Q}_r$ enforces the target-hand
joint and coupling constraints.

For Linker O6, we use a DexPilot-style~\cite{dexpilot} objective constructed from 15
geometric vectors: five wrist-to-fingertip vectors and ten pairwise
fingertip vectors. The optimizer solves for the six active O6 joints, while
the remaining joints follow the mechanical coupling and URDF mimic
constraints. For Wuji, the objective additionally incorporates
palm-to-PIP, palm-to-DIP, and palm-to-fingertip vectors, fingertip
directions, and thumb--finger pinch distances. Its weighting adapts to the
human grasp state, emphasizing whole-finger shape away from pinching and
fingertip position and direction near pinching. The resulting 20-DoF
configuration is solved using SLSQP with analytical kinematic gradients and
joint-limit constraints.

The tracked human wrist trajectory $T_t^w$ is mapped through
the calibrated human-to-robot arm transform to construct the relative
end-effector translation $\Delta x_{t,k}^r$ and relative rotation
$r_{t,k}^{r,6D}$, where the latter uses the 6-D rotation representation.

The optimized hand state is first converted to the native target-command
coordinates using a fixed embodiment-specific mapping $C_r$:
\begin{equation}
\begin{aligned}
u_t^r &= C_r(q_t^r),\\
h_{t,k}^r &= u_{t+k}^r-u_t^r,\\
a_{t+k}^r
&=
\left[
\Delta x_{t,k}^r,\,
r_{t,k}^{r,6D},\,
h_{t,k}^r
\right]^\top
\in\mathbb{R}^{9+d_r}.
\end{aligned}
\label{eq:robot-action}
\end{equation}
Here $h_{t,k}^r$ is a target-command delta rather than an absolute joint
configuration. For O6, $C_r$ maps the six optimized active-joint targets to
the native $0$--$250$ actuator command range, while the remaining joints are
handled by the mechanical coupling. The command dimensions are
$d_{\mathrm{O6}}=6$ and $d_{\mathrm{Wuji}}=20$.

During training, $q_t^r$ is obtained from retargeted human motion. During deployment, the measured robot state is converted to the same target-hand state representation and normalized using the corresponding training-set statistics.

\subsection{Grasp-Centered Interaction Representation}
\label{sec:pocket}

The interaction branch provides a compact description of the object relative
to the current grasp location. For thumb--index grasps, fingertip tracks are
smoothed with a causal three-frame median filter:
\begin{equation}
\begin{aligned}
\widetilde U_{t,j}^v
&=
\operatorname{med}
\left\{
U_{t-2,j}^v,U_{t-1,j}^v,U_{t,j}^v
\right\},\\
p_{t,j}^h
&=
\Gamma_t^h
\left(
\widetilde U_{t,j}^v
\right),\qquad
p_{t,j}^r=\bar U_{t,j}^r.
\end{aligned}
\label{eq:pocket-landmarks}
\end{equation}

For either domain $d\in\{h,r\}$, the Functional Grasp Pocket is the
image-space midpoint of the thumb and index fingertips:
\begin{equation}
g_t^d
=
\frac{
p_{t,\mathrm{thumb}}^d+
p_{t,\mathrm{index}}^d
}{2}.
\label{eq:grasp-pocket}
\end{equation}
Robot fingertips are projected individually before their midpoint is computed.

For object $i$, SAM-based tracking~\cite{sam2} provides the policy-image
centroid $(u_t^i,v_t^i)$ and mask area $A_t^i$. Short tracking failures are
handled using the last valid measurement, with confidence decay and a
memory-valid flag indicating whether the current state is directly observed
or temporally retained. With $g_t=(u_t^g,v_t^g)$, the Grasp-Pocket Interaction Representation is
\begin{equation}
z_t^i
=
\left[
\frac{u_t^i-u_t^g}{s_{\mathrm{ref}}},
\frac{v_t^i-v_t^g}{s_{\mathrm{ref}}},
\log\frac{A_t^i}{A_{\mathrm{ref}}^i},
c_t^i,
m_t^i
\right]^\top,
\label{eq:gpir}
\end{equation}
where $s_{\mathrm{ref}}$ is the image-coordinate normalization scale,
$A_{\mathrm{ref}}^i$ is the reference object area, $c_t^i$ denotes tracking
confidence, and $m_t^i$ indicates whether the object state is currently
observed or retained from memory. Multiple object states are
concatenated. GPIR provides a compact whole-object, image-plane representation
of grasp-relative alignment, scale, and tracking reliability for policy
conditioning.

\subsection{Policy Learning}
\label{sec:policy-learning}

The converted observation, target-hand state, and interaction state form the
policy input:
\begin{equation}
o_t^{r,\tau}
=
\left\{
\bar I_t,\,
q_t^r,\,
z_t
\right\},
\label{eq:policy-observation}
\end{equation}
where $q_t^r$ is always included and $z_t$ is included when interaction
grounding is enabled. Each hand--task pair is trained independently.

\begin{figure}[t]
    \centering
    \includegraphics[
        width=\columnwidth,
        trim=160bp 95bp 280bp 50bp,
        clip
    ]{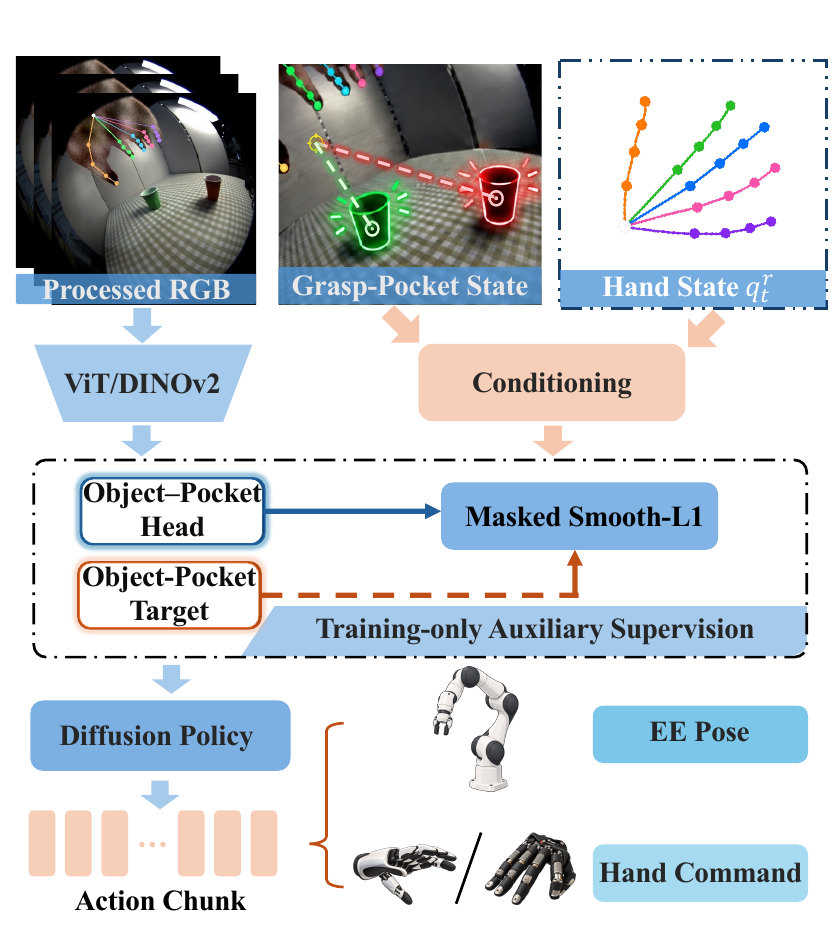}
    \caption{Target-specific policy learning and inference. Converted human
    observations condition a diffusion policy that predicts native commands
    for one hand--task pair. The human-data interface is shared across
    embodiments, whereas action construction, normalization, policy training,
    and checkpoints remain hand-specific.}
    \label{fig:policy}
\end{figure}

The policy architecture is shown in Fig.~\ref{fig:policy}. A DINOv2
ViT-B/14 Reg4 encoder~\cite{dinov2} processes $\bar I_t$. Its features are
combined with $q_t^r$ and, when enabled, $z_t$ to condition a 1-D U-Net. The
network predicts a 16-step action sequence under the Diffusion Policy
objective~\cite{diffusionpolicy}:
\begin{equation}
\pi_{\theta_{r,\tau}}
:
o_{t-h+1:t}^{r,\tau}
\mapsto
a_{t:t+H_a-1}^{r}.
\label{eq:policy}
\end{equation}

For configurations with auxiliary supervision, we regress the
Pocket-relative object position and the log-area:
\begin{equation}
y_t^{\mathrm{aux}}
=
\left[
\frac{u_t^i-u_t^g}{s_{\mathrm{ref}}},
\frac{v_t^i-v_t^g}{s_{\mathrm{ref}}},
\log\frac{A_t^i}{A_{\mathrm{ref}}^i}
\right]_{i=1}^{N}.
\label{eq:aux-target}
\end{equation}

The training objective is
\begin{equation}
\mathcal{L}
=
\mathcal{L}_{\mathrm{DP}}
+
\lambda_{\mathrm{aux}}
\mathcal{L}_{\mathrm{aux}},
\label{eq:training-objective}
\end{equation}
where $\mathcal{L}_{\mathrm{aux}}$ is a masked Smooth-L1 loss over valid
Pocket-relative object-state targets, and $\lambda_{\mathrm{aux}}=0$ when
auxiliary supervision is disabled. The auxiliary head is omitted during
deployment.

We train each policy with AdamW using a learning rate of $3\times10^{-4}$, a
batch size of 32, one-frame observation history, and U-Net channel widths of
256, 512, and 1024. The diffusion schedule uses 50 training timesteps and 16
DDIM inference steps. Five percent of episodes are reserved for validation,
with all appearance variants of an episode assigned to the same split. At deployment, the policy runs at 16~Hz and executes the first eight actions
of each predicted chunk before replanning from the latest observation.

\section{Experiments}

\subsection{Research Questions}

The experiments address three questions. \textbf{Q1:} Can one human demonstration corpus support independent policy learning for two dexterous hands? \textbf{Q2:} How do observation alignment and fused human-hand supervision affect performance, and how does wrist-centric observation compare with head-centric observation? \textbf{Q3:} When is grasp-centered interaction grounding useful, and how do reference coordinates and task geometry affect its performance?

\subsection{Experimental Setup}

\noindent\textbf{Target Robot Hands.}
We evaluate \method{} across two anthropomorphic dexterous hands with different kinematic structures. Both hands are equipped with calibrated wrist-mounted fisheye cameras:
\begin{itemize}
    
\item \textbf{Linker O6:} eleven-DoF anthropomorphic dexterous hand with six active DoFs. The thumb has two active and one passive DoFs, while each remaining finger has one active and one passive DoF.

\item \textbf{Wuji:} A fully-actuated hand with twenty active DoFs. Each finger has four active DoFs.
\end{itemize}

\noindent\textbf{Tasks.}
Both hands are evaluated on six real-world tasks spanning compliant grasping,
precise placement, object reorientation, coordinated finger contact, and
articulated-object interaction: \textbf{Sponge}, pick and place a compliant
sponge onto a target plate; \textbf{Croissant}, pick and place while
maintaining a stable grasp; \textbf{Cup}, insert a green cup into a red cup;
\textbf{Bottle}, reorient a horizontal bottle and place it upright;
\textbf{Egg Carton}, stabilize the carton and lift its lid; and
\textbf{Open Drawer}, engage the handle and pull the drawer open.

\begin{figure*}[t]
    \centering
    \includegraphics[
        width=0.95\textwidth,
        trim=10 140 10 120,
        clip
    ]{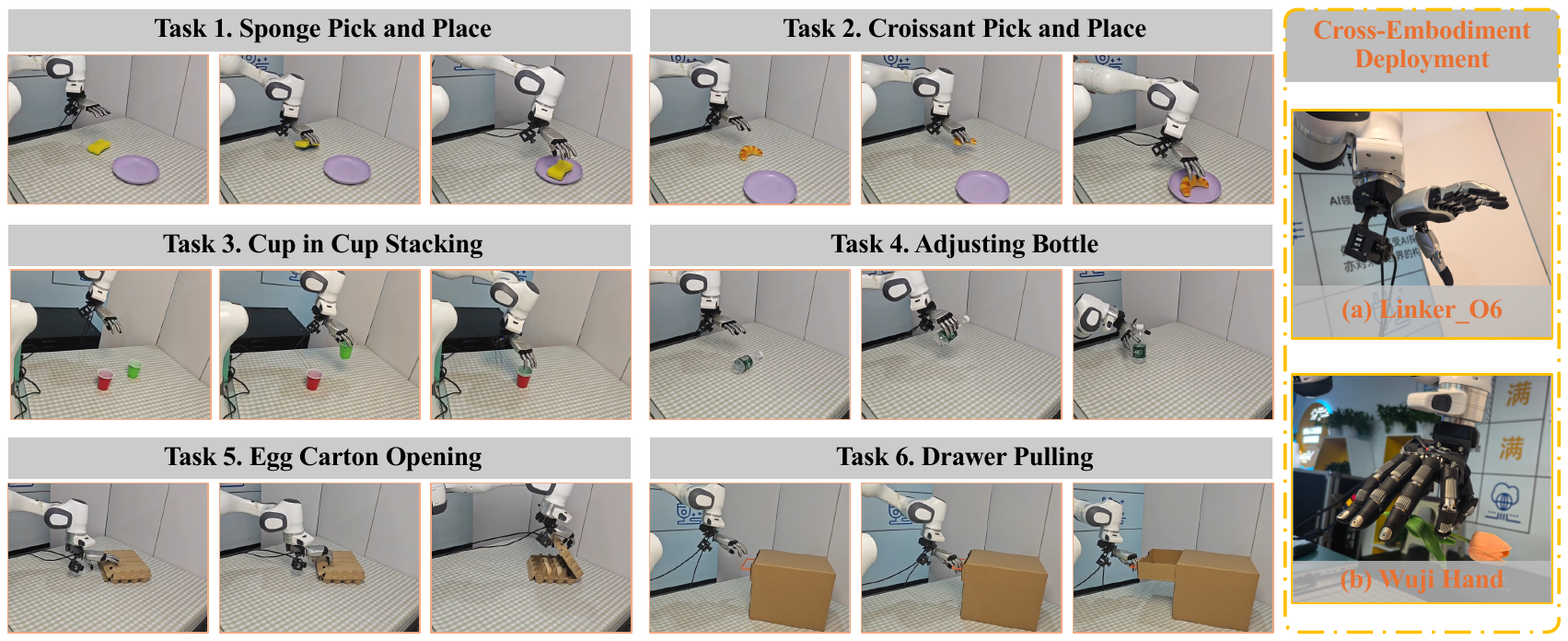}
    \caption{The six real-world manipulation tasks evaluated on the Linker O6
    and Wuji dexterous hands.}
    \label{fig:tasks}
\end{figure*}

\noindent\textbf{Shared Human Corpus.} 
The six tasks contain 1,899 quality-controlled base episodes and 556,624
frames. Each base episode produces three appearance-rendered variants. All
configurations use the same base episodes, episode-level split, sampled
variants, and optimizer budget. O6 and Wuji share episode groups but use
distinct actions, normalization, and checkpoints, with all variants of an
episode assigned to the same split.

\noindent\textbf{System Configurations.}
All configurations use the same embodiment-specific retargeter.
Human-Raw and Human-Visual use PICO-derived hand structures, whereas
Human-Structured and \method{} use the fused PICO/RGB structure.
Human-Visual adds hand-local appearance adaptation and the shared 21-point
structural rendering. Human-Structured further replaces PICO-only hand
supervision with fused PICO/RGB supervision, while retaining the same visual
alignment. Human-Structured has no explicit object state or auxiliary
object-regression head. \method{} additionally uses task-specific object
tracking, Pocket-relative object input, and the auxiliary objective.
Human-Absolute uses the same object processing and auxiliary supervision as
\method{} but references object position to the image center.

\noindent\textbf{Evaluation Protocol.}
Each method--task pair receives 20 physical trials with fixed reset ranges; a success must satisfy its terminal criterion for 2~s. Configurations share the base demonstrations, grouped split, backbone, optimization budget, and physical protocol. Wilson 95\% intervals describe trial uncertainty for these fixed configurations.  Cross-task pooled counts are benchmark summaries, not estimates for unseen-task populations.

\subsection{Main System-Level Results}
\label{sec:main-results}

Table~\ref{tab:main_results} addresses Q1. Both hands use the same human demonstration sources, while hand supervision and policies are independently constructed for each embodiment. The four rows are separately trained system configurations.

\begin{table*}[t]
\centering
\caption{Real-world task success (successful trials / 20). Each hand--task pair uses independently retargeted supervision and a separate policy; the final row uses task-specific object--pocket observations on all six tasks.}
\label{tab:main_results}
\resizebox{\textwidth}{!}{
\begin{tabular}{@{}lcccccccccccc@{}}
\toprule
\multirow{2}{*}{Method}
& \multicolumn{6}{c}{Wuji}
& \multicolumn{6}{c}{Linker O6} \\
\cmidrule(lr){2-7}
\cmidrule(lr){8-13}
& Sponge & Croissant & Cup & Bottle & \shortstack{Egg\\Carton} &
\shortstack{Open\\Drawer}
& Sponge & Croissant & Cup & Bottle & \shortstack{Egg\\Carton} &
\shortstack{Open\\Drawer} \\
\midrule

Human-Raw
& 5/20 & 3/20 & 2/20 & 0/20 & 2/20 & 7/20
& 10/20 & 6/20 & 5/20 & 6/20 & 12/20 & 5/20 \\

Human-Visual
& 12/20 & 8/20 & 7/20 & 12/20 & 13/20 & 13/20
& 14/20 & 8/20 & 11/20 & 16/20 & 16/20 & 14/20 \\

Human-Structured
& 12/20 & 10/20 & 7/20 &
15/20 & \textbf{14/20} & \textbf{16/20}
& 15/20 & 9/20 & 11/20 &
17/20 & \textbf{20/20} & \textbf{17/20} \\

\textbf{\method{}}
& \textbf{19/20} & \textbf{17/20} & \textbf{17/20}
& \textbf{16/20} & 13/20 & 15/20
& \textbf{20/20} & \textbf{18/20} & \textbf{16/20}
& \textbf{19/20} & 19/20 & \textbf{17/20} \\

\bottomrule
\end{tabular}
}
\end{table*}

Human-Raw obtains $19/120$ on Wuji (15.8\%) and $44/120$ on O6 (36.7\%). \method{} reaches $97/120$ on Wuji (80.8\%, Wilson 95\% CI 72.9--86.9\%) and $109/120$ on O6 (90.8\%, 84.3--94.8\%). Across both hands, the pooled success rates for Human-Raw, Human-Visual, Human-Structured, and \method{} are 26.3\%, 60.0\%, 67.9\%, and 85.8\%, respectively.

These results support the reuse of one human demonstration corpus for independent policy learning on two calibrated dexterous hands. The successive configuration comparisons provide system-level evidence for the integrated conversion framework; they do not isolate the causal effect of any individual component.

\subsection{Observation and Human-Motion Supervision Analyses}
\label{sec:observation-motion-analysis}

The Raw-to-Visual and Visual-to-Structured comparisons characterize the
performance associated with observation alignment and human-motion
supervision. All configurations use the same embodiment-specific retargeter;
the Structured configuration instead replaces PICO-only hand structures with
fused PICO/RGB supervision. Human-Raw achieves 26.3\% pooled success,
Human-Visual achieves 60.0\%, and Human-Structured achieves 67.9\%.
These are separately trained system configurations, with embodiment-specific
retargeting fixed across conditions.

We evaluate a matched O6–Croissant comparison using synchronized head- and wrist-camera observations while fixing action supervision, architecture, optimization, and physical evaluation. This viewpoint study is conducted as a separate matched diagnostic rather than as the full system-level condition reported in Table I; therefore, its absolute success rate should not be directly compared with the Human2Dex result in Table I. First-Attempt Grasp denotes a stable first closure without reopening or regrasping. It is used here as a secondary viewpoint diagnostic, not as an independent measure of GPIR. Both conditions use the Human-Visual configuration with identical action supervision; only the human demonstration camera differs.

Wrist-view demonstrations achieve 6/20 final successes and 4/20 first-attempt grasps, whereas head-view demonstrations achieve 0/20 for both measures. This single-hand, single-task comparison supports wrist-centric observation as a useful design choice for the evaluated eye-in-hand setting.

\begin{figure}[t]
    \centering
    \includegraphics[
        width=\columnwidth,
        trim=130bp 140bp 210bp 160bp,
        clip
    ]{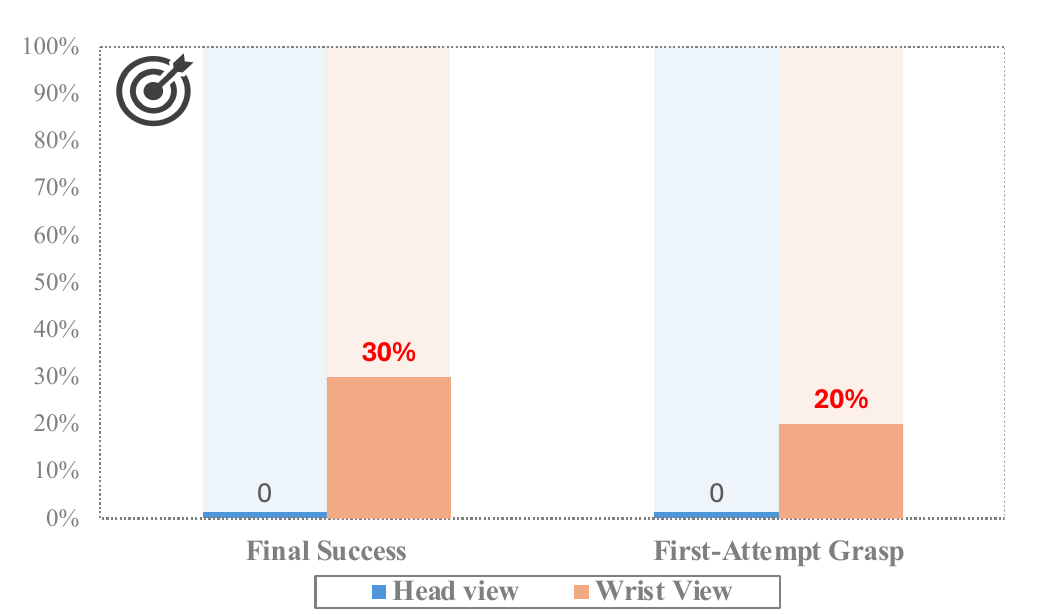}
    \caption{Matched head- versus wrist-view comparison on O6–Croissant. 20 trials per condition, wrist-view demonstrations achieve 30\% final success and 20\% first-attempt grasp success, compared with 0\% for both metrics using head-view demonstrations.}
    \label{fig:view-comparison}
\end{figure}


\subsection{Interaction-Grounding Analysis}
\label{sec:interaction-analysis}

\noindent\textbf{Integrated grounding.} The Structured-to-\method{} comparison evaluates the complete interaction-grounding configuration. This transition adds object tracking, Pocket-relative object input, and the auxiliary prediction objective together.

\noindent\textbf{Reference-frame comparison.} We compare Human-Absolute and \method{} on Linker O6 for Croissant and Cup. The two configurations use the same base episodes, object processing, architecture, optimization, auxiliary target, and physical evaluation protocol. They differ only in whether object position is referenced to the image center or the Functional Grasp Pocket.

Human-Absolute reaches $17/20$ on Croissant, compared with $18/20$ for \method{}. On Cup, Human-Absolute reaches $9/20$, compared with $16/20$ for \method{}. Image-centered conditioning is competitive for Croissant, whereas Cup shows a larger improvement with the Pocket-relative reference in this evaluation.

\noindent\textbf{Task-wise behavior.} Across the six tasks, the complete interface improves performance most on Sponge, Croissant, Cup, and Bottle, while the gains are smaller or absent on Egg Carton and Open Drawer. This task dependence is consistent with the scope of the current object-level grounding. Sponge, Croissant, Cup, and Bottle primarily require the hand to align with and grasp the object as a whole, for which an object-to-pocket relation provides a meaningful geometric cue. In contrast, Egg Carton and Open Drawer require interaction with localized functional regions---the lid edge and drawer handle---whose locations are not necessarily aligned with the centroid of the tracked object. These results suggest that the current GPIR is most effective for grasp-centric tasks, while tasks dominated by localized contact may benefit from part-level functional grounding.

\begin{figure}[t]
    \centering
    \includegraphics[
        width=\columnwidth,
        trim=180bp 140bp 145bp 100bp,
        clip
    ]{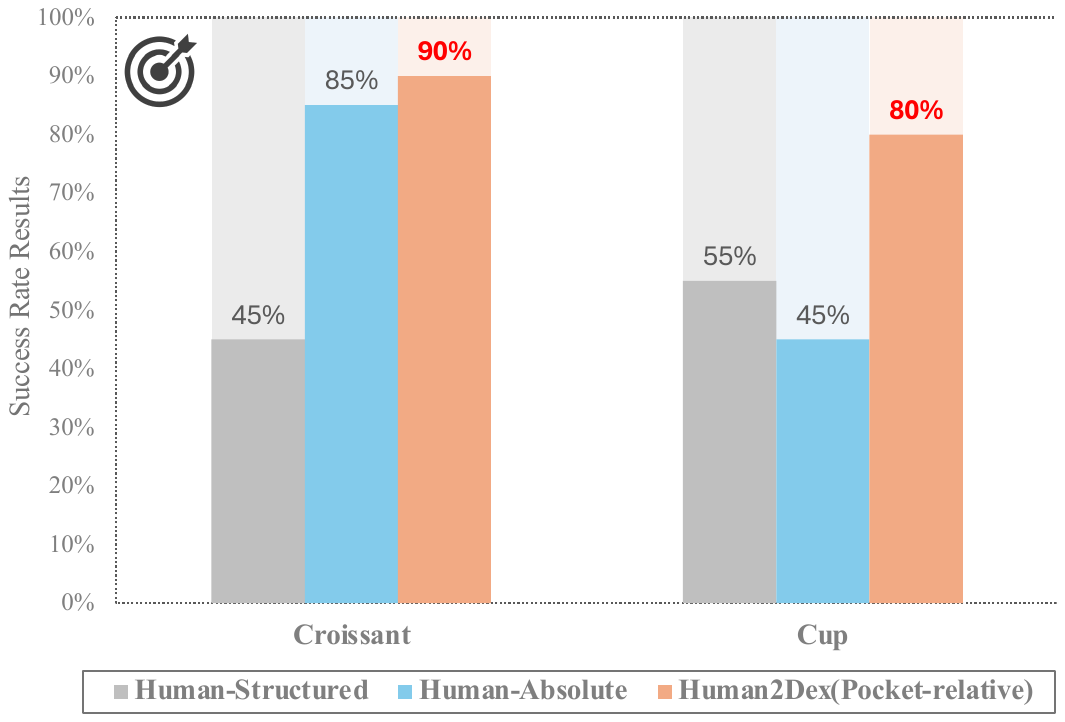}
    \caption{Functional-reference comparison on Linker O6. Bars show task success rate over 20 physical trials. Human-Absolute and Human2Dex differ only in whether object position is referenced to the image center or the Functional Grasp Pocket.}
    \label{fig:functional-reference}
\end{figure}





\section{Discussion and Limitations}

The main result is that one human demonstration source can support independent policy learning on two calibrated dexterous hands. This reuse is achieved by separating shared source semantics from target-hand execution: human observations and motion representations are shared, while retargeting, executable action labels, normalization, and policy training are constructed for each hand--task pair. The resulting interface avoids recollecting task demonstrations for each target hand while preserving the control structure required by each embodiment.

The reference-frame comparison shows that object localization and interaction referencing are distinct design choices. In the controlled comparison, image-centered conditioning is competitive on Croissant, whereas Pocket-relative conditioning performs substantially better on Cup, where grasp alignment and insertion are more precise. These results suggest that the value of a grasp-centered reference depends on how directly the object--hand relation determines task success.

The task-wise results further clarify the role of interaction grounding. The Structured-to-\method{} comparison adds object tracking, Pocket-relative object input, and the auxiliary prediction objective together, and therefore evaluates the integrated grounding configuration. Improvements on Sponge, Croissant, Cup, and Bottle are consistent with tasks that require alignment with and grasping of an object as a whole. Egg Carton and Open Drawer depend more on localized regions such as a lid edge or drawer handle, where a whole-object representation provides a less direct cue. These results motivate part-level or affordance-level interaction representations for localized contact tasks.

The present study covers two calibrated anthropomorphic hands and six real-world tasks, with a separate policy for every hand--task pair. The evaluation therefore establishes reuse across the tested embodiments, while broader transfer to arbitrary morphologies and shared action spaces remains open. GPIR currently uses a specified task object and reliable image-space tracking, and its whole-object representation does not explicitly identify localized contact regions. The viewpoint comparison is limited to one hand and one task. In addition, the absence of a robot-demonstration baseline leaves direct sample-efficiency comparisons between human and robot data for future work.

\section{Conclusion}

Human2Dex establishes a reusable wrist-centric human demonstration interface for dexterous robot learning across multiple embodiments. It shares human demonstrations and their observation and motion semantics, then converts them into embodiment-specific visual observations and executable action supervision for each target hand, with GPIR providing a grasp-centered object--hand interaction cue. Using one human demonstration corpus, Human2Dex supports independent policy learning for Linker O6 and Wuji and achieves 109/120 and 97/120 successes across six real-world tasks, respectively. The reference-frame and task-wise analyses further suggest that whole-object grasp-relative grounding is most beneficial when task success depends strongly on grasp alignment, whereas localized interactions may require finer-grained representations.
\bibliographystyle{IEEEtran}
\bibliography{refs}

\end{document}